# Multi-Task Attentive Residual Networks for Argument Mining

Andrea Galassi, Marco Lippi, and Paolo Torroni

*Abstract*—**We explore the use of residual networks and neural attention for multiple argument mining tasks. We propose a residual architecture that exploits attention, multi-task learning, and makes use of ensemble, without any assumption on document or argument structure. We present an extensive experimental evaluation on five different corpora of user-generated comments, scientific publications, and persuasive essays. Our results show that our approach is a strong competitor against state-of-the-art architectures with a higher computational footprint or corpus-specific design, representing an interesting compromise between generality, performance accuracy and reduced model size.**

## I. Introduction

Argument mining (AM) is an area of natural language processing (NLP) defined by a variety of tasks, aiming to extract and structure arguments from unstructured text [1]. Some are argument detection, stance classification and topic-based argumentative content retrieval [2]. The problem we address in this work is to assemble the structure of the argumentation behind a given input document. This problem can be broken down into multiple tasks, such as the detection of argument components, as well as the classification of links between them. The latter is known to be a challenging task, whose outcome may be a complicated graph.

There are many possible definitions of argument. According to Walton [3], an argument is made of three components: (i) a *claim*, or assertion, about a given topic; (ii) a set of *premises* supporting the claim; and (iii) the inference between the premises and the claim. Relations between arguments, or argument components, typically consist of either *support* or *attack* links. Argument components and relations may be implicit, which contributes to the difficulty of the task at hand. Moreover, not all argument definitions fit all genres. In fact, AM approaches are very often tailored to specific corpora or genres [4], [5], with solutions that are seldom general enough to be directly applicable to different data sets. Indeed, many AM systems build upon sets of handcrafted features which encode information about the underlying argument model, genre or topic of interest, and make assumptions on the argumentative structure of the input document, thus constraining the resulting argument graph.

On the other end of the spectrum, we find increasingly many solutions that do not rely on feature engineering, but on huge neural architectures with millions of trainable parameters. These models are usually very accurate, but also very expensive, especially in terms of the carbon footprint resulting from the huge energy cost of training and fine-tuning [6]. So much so that a significant part of the NLP community is now promoting the vision of a Green AI, whereby more effort must be spent on simpler and efficient solutions, suited for low-resources settings [7], [8], [9], [10].

In the last years, the availability and diversity of AM corpora has considerably increased [2], [11], [12]. However, most AM models are tested only on a few popular benchmarks, typically neglecting less known datasets, and only reporting on positive results. This phenomenon, which is not limited to AM research but has been observed in other communities as well, has been often criticized because it may hinder the development of new ideas [13] and promote the development of models that generalize poorly to the real world [14].

This work presents a general-purpose, domain-agnostic neural architecture that does not rely on genre-specific or topic-dependent features, and its evaluation on five different datasets. The architecture is smaller than state-of-the-art models by various orders of magnitude. It exploits neural attention and multi-task learning, jointly addressing the problems of identifying the category of argument components, and predicting their relations. Experimental results conducted on a variety of different corpora show that the model is robust, can be applied to many different domains, and achieves good performance across the considered data sets. Our main contributions are:

- A novel approach to AM, which extends our previous work [15] by introducing an attention module and using ensemble learning. The model jointly performs multiple AM tasks, and does not rely on ad-hoc features or rich

Manuscript received 7 November 2021; revised 19 September 2022 and 15 April 2023; accepted 5 May 2023. Date of publication 10 May 2023; date of current version 17 May 2023. This work was supported in part by the European Union's Justice Programme, Project ADELE, under Grant 101007420, in part by the European Union's Horizon 2020 Programme, Project StairwAI, under Grant 101017142, and in part by the European Commission's NextGeneration EU Programme, PNRR - M4C2 - Investimento 1.3, Partenariato Esteso PE00000013 - FAIR - Future Artificial Intelligence Research - Spoke 8 Pervasive AI. The associate editor coordinating the review of this manuscript and approving it for publication was Dr. Preslav Nakov. *(Corresponding author: Andrea Galassi.)*

Andrea Galassi and Paolo Torroni are with the Department of Computer Science and Engineering (DISI), University of Bologna, 40126 Bologna, Italy (e-mail: a.galassi@unibo.it; paolo.torroni@unibo.it).

Marco Lippi is with the Department of Sciences and Methods for Engineering (DISMI), University of Modena and Reggio Emilia, 42122 Modena, Italy (e-mail: marco.lippi@unimore.it).

contextual information, but only on GloVe embeddings and on a widely applicable notion of distance.

- A model with a much smaller computational footprint than state-of-the-art neural approaches.
- An analytical evaluation of the contribution of each added module through an ablation study and a validation of our model on a challenging corpus.
- A set of experiments to assess generality, whereby we test our approach on four different corpora with various domains, writing style, formatting, length, and annotation model. To the best of our knowledge, we are the first to validate a new AM method on as many corpora.
- A negative result on a fifth corpus, which highlights the limitations of our approach, as suggested in [16].

With respect to our previous work [15], this article extends the neural architecture with attention and ensemble learning, and presents a more extensive experimental evaluation. All the code used in our experiments is publicly available.[1]

The article is organized as follows. Section II presents background and related work. Section III introduces our architectures. Section IV describes the data used for evaluation. Sections V and VI illustrate the experimental setting and discuss results. Section VII concludes.

## II. Background

The adoption of deep learning approaches in AM is relatively recent, compared to other areas of NLP. That is probably a consequence of a lack of large AM corpora, considering the complexity and peculiarities of the tasks at hand. Indeed, the annotation of large corpora for AM system evaluation and training proved to be challenging, as demonstrated by relatively low Inter-Annotator Agreement (IAA) indicators and several unsatisfactory attempts at crowdsourcing annotations. That is especially true for some genres like user-generated content [17]. Reasons for that are the nature of the task, which is intellectually demanding, and the lack of a unified argument model, as "arguments" may take very different shapes in different genres, also leading to a trade-off between the expressiveness of the argument model and the complexity of the annotation process and availability of relevant data points, often resolved in favor or simple argument models [1]. Earlier research mainly focused on the definition of features for specific genres or even for specific corpora. The differences between corpora, both regarding the domain and the theoretical framework followed during the annotation process, forced researchers to test a model on the same corpora on which it was trained, and to the best of our knowledge, transfer learning approaches have not seen wide experimentation. These two elements lead to the common practice to define a method or a model and validate it only on a single corpus or on a few corpora [1].

### A. Multi-Task Learning and Joint Learning for am

Since AM includes many subtasks that are strongly interrelated, a recent trend of this research field is to address many of them at the same time using multi-task or joint learning techniques. The aim of such approaches is to transfer knowledge from the auxiliary tasks to the main one, or to obtain coherent results on multiple tasks performed at once.

Stab and Gurevych [4] jointly address component classification and link prediction on persuasive essays, using Integer Linear Programming and a rich set of specific features, such as lexical, structural, and contextual information. Various neural architectures are tested in [18], including the deep biLSTM multi-task learning (MTL) setting of [19], using sub-tasks as auxiliary tasks. They conclude that neural networks can outperform feature-based techniques in argument mining tasks. Schulz et al. [20] investigate MTL settings addressing component detection on five datasets as five different tasks. Their architecture is composed of a CRF layer on top of a biLSTM, whose recurrent layers are shared across the tasks. They obtain positive results, and the MTL setting shows to be beneficial especially for small datasets, even if the auxiliary AM tasks involve different domains and even different component classes. Lauscher et al. [21] analyze an MTL setting where rhetorical classification tasks are performed along with component detection. They use a hierarchical attention-based model to perform both word-level and sentence-level tasks with the same neural architecture. The results show improvements in the rhetorical tasks, but not in AM. Accuosto and Saggion [22] experiment with MTL and sequential transfer learning, improving performance on AM through discourse parsing tasks.

In [23], a structured learning framework based on factor graphs is used to jointly classify all the propositions in a document and determine which ones are linked together. The models heavily rely on a priori knowledge, encoded as factors and constraints, designed to enforce adherence to the desired argumentation structure, according to the argument model and domain characteristics. The authors discuss experiments with six different models, which differ by complexity and by how they model the factors, using RNNs and SVMs. Their best result is obtained by using the same set of features used in [4], resulting in a total feature size of around 7,000 for propositions and 2,100 for links. Finally, another approach based on factor graph is DRAIL [24], a neuro-symbolic framework that allows to specify the structure and the constraints of the graphs through first-order logic clauses.

### B. Neural Attention for AM

Neural attention is a mechanism widely used in NLP to improve performance and interpretability of neural networks, and it is the core of many NLP architectures like RNNsearch [25], Pointer Networks [26], and Transformer [27]. Given an input sequence, and possibly a query element, attention consists in the computation of a set of weights that represent the importance of each element of the sequence, which can be further used to create a compact representation of such an input. There are many different ways to compute such weights. A taxonomy of attention models is proposed in our survey [28].

Among the AM systems that use neural attention, the one used in [29] integrate hierarchical attention and biGRU for the analysis of the quality of the argument, the one in [30] use attention to integrate sentiment lexicon, while in other works [31], [32], [33] attention modules are stacked on top of recurrent layers. The use of Pointer Networks for AM has also been investigated [34]. Biaffine attention has been used by Morio et al. [35] along with task-specific parametrization (TSP-PLBA) and a mixture of symbolic and sub-symbolic input features. Chen et al. [36] address the task of inferring the agreement between sentences using fine-grained co-attention between the two sentences.

[1] [Online]. Available: https://github.com/AGalassi/StructurePrediction18.

Transformer-based approaches in AM use language representation models such as BERT [37] and ELMO [38] to create contextualized word embeddings. Specifically, Reimers et al. [39] address component classification and argument clustering, a related task whose aim is to identify similar arguments. Similarly, Lugini and Litman [40] use BERT embeddings alongside other contextual information to perform component classification, and Wang et al. [41] use them to train a different model for each type of component. Trautmann et al. [42] use pre-trained BERT models to perform word-level classification of the stance of components regarding a given topic, while Poudyal et al. [43] use RoBERTa [44], an improved version of the original BERT. BERT is also used by Opitz [45], who formulates relation classification as a plausibility ranking task by exploiting hypothetical discourse contexts. Bao et al. [46] propose a neural transition-based model which incrementally builds an argumentation graph, using a combination of fine-tuned BERT embeddings and other symbolic features as input. More recently, Srivastava et al. [47] use BERT to classify arguments, then rely on the trained weights and self-attention to predict links.

Mayer et al. [48] present and conduct extensive experimentation on the AbstRCT corpus, addressing four AM subtasks with a pipeline scheme. They analyze the impact of various BERT models, which are pre-trained on other corpora and then fine-tuned on the corpus at hand. Segmentation and component classification are performed as sequence tagging with BIO scheme. Link prediction and relation classification follow, taking into account all the pairs of components obtained in the first step and classifying their relations as attack, support, or non-existing. Their architecture is based on bi-directional transformers followed by a softmax layer and various encoders. Their approach is completely distance-independent, but since they compare every possible pair of components, the size of the dataset grows quadratically with the number of components in the document, which makes it hardly scalable to large documents. Another approach, consisting of predicting at most one related component for each component, and then classifying their relation, has been tested but yields worse results. The architectures that yield the best results are BioBERT [49], which is pre-trained on a large-scale biomedical corpus, SciBERT [50], which is pre-trained on scientific articles of various nature, and RoBERTa.

Looking outside the context of AM, BERT is a very popular model across all the NLP tasks [51], but it is also very resource-demanding, consisting of more than 110 millions parameters in its base implementation. For this reason, there is an active effort in assessing when its use is really necessary [8].

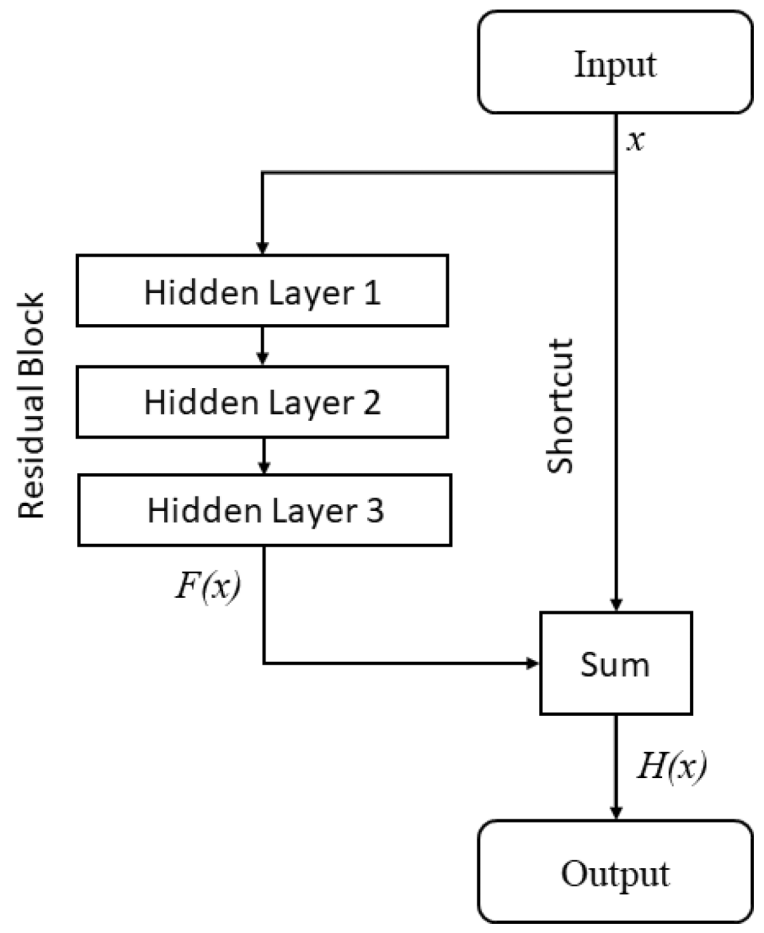

Fig. 1. Block scheme of a residual network.

### C. Residual Networks

Residual networks [52] are a family of deep neural networks that achieved outstanding results in many machine learning tasks across many different domains related to NLP [27], [53], [54]. The core idea behind residual networks is to create shortcuts that link neurons belonging to distant layers (see Fig. 1), whereas standard feed-forward networks typically link neurons belonging to subsequent layers only. This kind of architecture usually results in a more efficient training phase, allowing to train networks with considerably more layers, reducing the overall computational footprint. A similar principle is followed also in the design of dense and highway networks [55], [56]. The intuition behind residual networks is that if a function $H(x)$ can be approximated by multiple non-linear layers, then they can also approximate its residual function $F(x) = H(x) - x$. It is therefore possible to obtain the original function simply adding the residual value: $H(x) = F(x) + x$.

## III. Model

The architecture we propose makes use of the dense residual network model, along with a Long Short-Term Memory (LSTM) network [57], and an attention module [28]. The network is trained to jointly perform three argument mining sub-tasks: argument component classification, link prediction, and relation classification.

More specifically, our approach operates on sentence pairs, does not rely on document-level global optimization, and does not enforce model constraints induced, for example, by domain- or genre-specific background knowledge. This makes our approach amenable to a possible integration within more complex and sophisticated systems.

We performed model selection and hyper-parameter tuning on a single corpus (CDCP, see Section IV) and we collected results on validation data in order to tune the whole architecture. There are two reasons for this choice: on the one hand, we aim to show the robustness of the approach across different corpora, while on the other hand we believe it is important to limit the footprint of these experiments – an issue that is receiving a growing attention in the community [6].

### A. Model Description

In order to achieve a general method which may be applicable in any domain, our approach does not rely on a specific argument model, but rather it reasons in terms of abstract entities, such as argumentative components and links among them. We instantiate such abstract entities into concrete categories given by annotations, such as claims and premises, supports and attacks, as soon as we apply the method to a specific corpus whose annotations follow a concrete argument model.

The detection of argumentative content in text is one typical stage of AM systems [1]. Other works only focus on AM tasks that assume that argumentative components and their boundaries are already identified in the data. Such is the case with Niculae et al. [23], whose CDCP dataset only consists of argumentative elements, and with others [31], [40], [58] who simply ignore the non-argumentative elements of the input text. Accordingly, we define a *document* $D$ as a sequence of *argumentative components* and disregard the rest of the input text. An argumentative component in turn is a sequence of *tokens*, i.e., words and punctuation marks, representing an argument, or part thereof. The labeling of components is induced by the chosen argument model. Such a labeling associates each component with the corresponding category $C$ of the argument component it contains. For this reason, we will use the terms component, sentence, and proposition as equivalent, and implying them as being argumentative by assumption.

Given two argumentative components $a$ and $b$ belonging to the same document, we represent a directed relation from the former (*source*) to the latter (*target*) as $a \rightarrow b$. Reflexive relations ($a \rightarrow a$) are not allowed.[2] Any pair of components is characterized by four labels: the types of the two components ($C_a$ and $C_b$), the Boolean *link label* $L_{a\rightarrow b}$, and *relation (type) label* ($R_{a\rightarrow b}$). The link label indicates the presence of a link, and is therefore *true* if there exists a directed link from $a$ to $b$, and *false* otherwise. The relation label instead contains information on the nature of the link connecting $a$ and $b$. It represents the relationship between the two components, according to the links that connect $a$ to $b$ or $b$ to $a$. Its domain is composed, according to the underlying argument model, not only by all the possible link types, but also by their opposite types (e.g., *attack* and *attackedBy*), as well as by a special category, *None*, meaning no link in either direction. One reason to introduce opposite relation types is to mitigate the unbalance caused by limited amount of instances each relation type typically has, if compared with the number of instances belonging to the *None* class. Likewise, we speculate that the introduction of additional labels may contribute positively to the optimization process. We shall remark that opposite relation labels are exploited during training, but they are discarded in the test phase, where they are simply substituted with the *None* label, consistently with previous work. We use a multi-objective learning setting where multiple tasks are performed jointly for each possible input pair of components $(a, b)$ belonging to the same document $D$. Our main focus is the identification of the link label $L_{a\rightarrow b}$ for each possible input pair of propositions $(a, b)$ belonging to the same document $D$. Our first objective is thus a *link prediction* task, which can be considered as a sub-task of argument structure prediction. A second objective is the *classification* of the two components,[3] and our final objective is the classification of the relationship between such components, i.e., the prediction of labels $C_a$, $C_b$, $R_{a\rightarrow b}$. A common issue in the classification of pairs of document components is the fact that pairs grow quadratically with the number of components, causing a large imbalance against the negative class [43], [48]. One way of dealing with that issue is to limit the possible pairs by setting a maximum distance, thus obtaining a number of pairs proportional to the number of components. Such a distance is a hyper-parameter, and as such it may be empirically determined [43].

### B. Embeddings and Features

Faithful to the main purpose of this work, of evaluating the effectiveness of deep residual networks and attention for AM without resorting to domain- or genre-specific information, our system relies on a minimal set of widely applicable features.

Words are encoded using pre-trained GloVe embeddings [59] of size 300. Since punctuation may play a key role in the semantic of the sentences [60], we have decided to keep punctuation tokens as well. Input sequences are zero-padded to the length of the longest sequence in the datasets (henceforth $T$). Out-of-vocabulary terms are handled by creating random embeddings.

In our previous work, we empirically assessed how the distance between two components may be a relevant feature for AM in the CDCP corpus [15]. The same observation has been recently made also with reference to other corpora [18], [43], [48]. Similarly to what has been done in [18], we define the number of argumentative components separating source and target as *argumentative distance*, using the positive sign when the source precedes the target, and the negative sign otherwise. Inspired by works in other domains [61], [62], [63], we encode such a scalar number in a 10-bit array, using the first 5 bits for those cases where the source precedes the target, and the other 5 bits for the opposite case. The number of consecutive "1" values encodes the value of the distance, with a maximum value of 5. For example, if the argumentative distance is $-3$, the encoding is 00111 00000; if the argumentative distance is 2, the encoding is 00000 11000.

### C. The ResArg Architecture

We use our own previous system [15] as a baseline. We refer to it as ResArg. Its architecture, depicted in Fig. 2(a), is based on residual networks [52] and comprises the following macro blocks:

- two deep embedders, one for sources and one for targets, that manipulate token embeddings;
- a dense encoding layer that reduces the dimensionality of the features;
- a biLSTM that processes the sequences;

[2] We will partially consider reflexive relations for the UKP dataset for a specific reason explained in Section V.

[3] Since we examine only argumentative propositions, we do not consider the non-argumentative class for component classification.

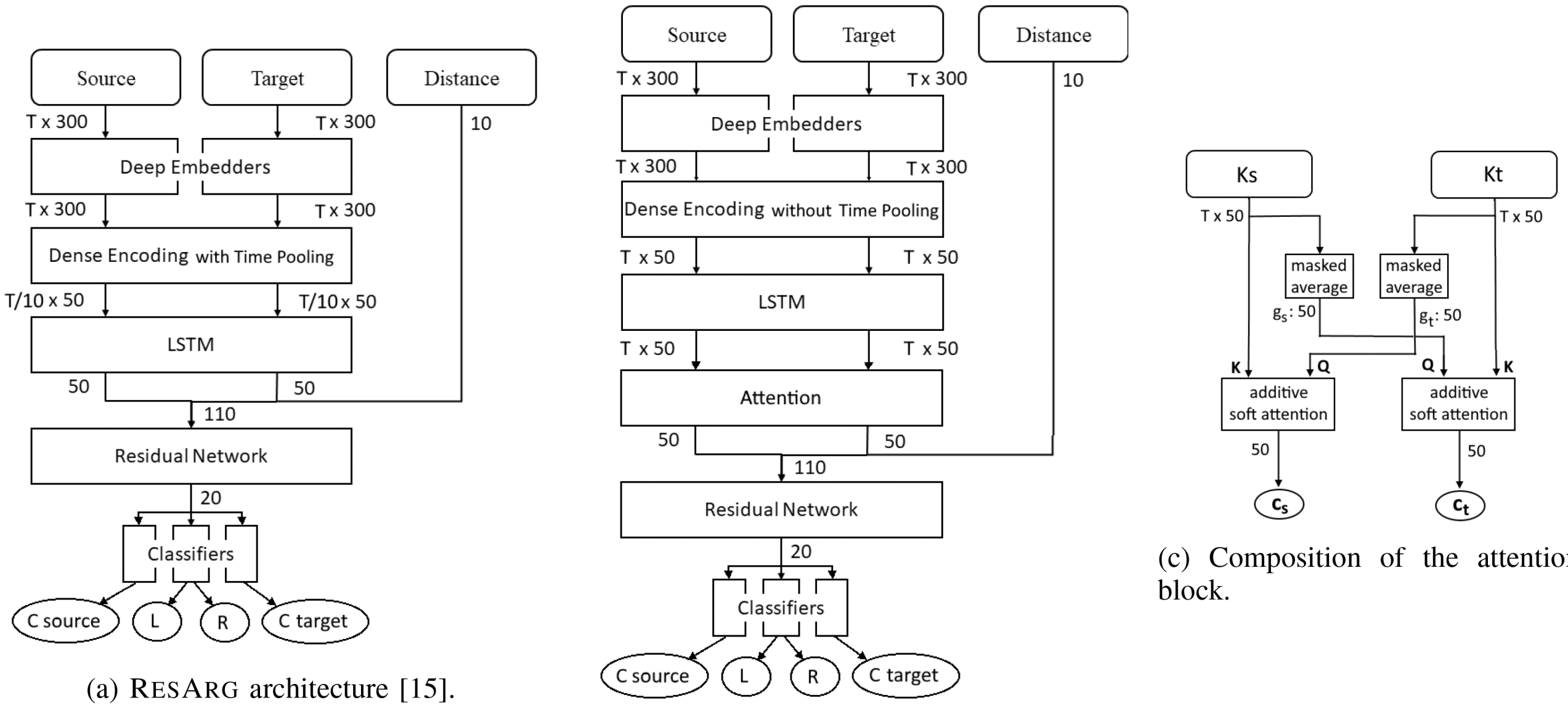

(a) RESARG architecture [15].

(b) RESATTARG architecture.

(c) Composition of the attention block.

Fig. 2. Block diagram of our previous architecture (a), our novel architecture (b), and the attention block we use (c). The figure shows, next to each arrow, the dimensionality of the data involved, so as to clarify the size of the inputs and the outputs of each block. $T$ is the temporal size of the data.

- a residual network;
- the final-stage classifiers.

The purpose of the deep embedders is to fine-tune the pre-trained embeddings, a common procedure in deep learning-based NLP solutions [64] whose usefulness was confirmed by preliminary experiments. Each embedder is composed of a single residual block consisting of four pre-activated time-distributed dense layers. Accordingly, each layer applies the same transformation to each embedding, regardless of their position inside the sentence. All the layers have 50 neurons, except for the last one, which has 300 neurons.

The dense encoding layer is necessary to reduce the parameters in the following biLSTM, thus reducing the time needed for training, and limiting overfitting. It applies a time-distributed dense layer, which reduces the embedding size to 50, and a time average-pooling layer [65], which reduces the sequence size by a factor of 10. The resulting sequences are then given as input to the same biLSTM, producing a single representation of size 50 for each component.

Source and target are processed in parallel in the first three blocks, then concatenated together, along with the encoding of the distance, and given as input to the final residual network. The first level of the final residual network is a dense encoding layer with 20 neurons, while the residual block is composed of a layer with 5 neurons and one with 20 neurons. The outputs of the first and the last layers of the residual networks are summed up and provided as input to the classifiers.

The final stage of RESARG are three independent softmax classifiers used to predict the source, the target, and the relation labels. Each classifier, which predicts a label for a dedicated task, contributes simultaneously to our learning model. The link classifier is obtained by summing the relevant scores produced by the relation classifier, aggregating the probability assigned to the relation labels into a single link label.

All the dense layers use the rectifier activation function [66], and they randomly initialize weights with He initialization [67]. The application of all non-linear functions is preceded by batch-normalization layers [68] and by dropout layers [69], with probability $p = 0.1$. The resulting architecture has about 130,000 trainable parameters.

### *D. The RESATTARG Architecture*

Motivated by the remarkable results obtained by attention-based architectures in NLP tasks, we have extended RESARG by including a neural attention block after the bi-LSTM module. To better exploit the new attention module, we removed the time pooling layer from the dense encoding block, so as to avoid loss of information along the temporal axis, and to maintain the whole output sequence from the LSTM. Therefore, in this new model, the input and the output of the LSTM module have size ($T$, 50). The resulting architecture, named RESATTARG, is depicted in Fig. 2(b).

The attention module is implemented as *coarse-grained parallel co-attention* [28], to consider both components at the same time while computing attention on each of them, and its structure is illustrated in Fig. 2(c). Our method consists of exploiting the average embedding of one proposition as a query element while computing attention on the other, similarly to what has been done in [70]. Specifically, calling $\boldsymbol{K_s}$ and $\boldsymbol{K_t}$ the outputs of the bi-LSTM obtained from, respectively, the processing of the source and the target propositions, we compute the (masked) average of $\boldsymbol{K_t}$, obtaining a single embedding $\boldsymbol{g_t}$ of size 50 (1). This embedding is used as query element to compute *additive soft attention* [28] on $\boldsymbol{K_s}$ (2), obtaining a set of attention weights $a_{si}$ that represent the relevance of an element (3), and then a single source context vector $\boldsymbol{c_s}$ of size 50 (4). The details of this process are described in the following Equations, where the matrices $\boldsymbol{W_1}$, $\boldsymbol{W_2}$ and the vectors $\boldsymbol{b}$, $\boldsymbol{w_3}$ are learnable parameters.

$$\boldsymbol{g_t} = masked_avg(\boldsymbol{K_t}) \quad (1)$$

$$\boldsymbol{e_s} = \boldsymbol{w_3}^T \; relu(\boldsymbol{W_1 K_s} + \boldsymbol{W_2} \, \boldsymbol{g_t} + \boldsymbol{b}) \quad (2)$$

$$\boldsymbol{a_s} = softmax(\boldsymbol{e_s}) \quad (3)$$

$$\boldsymbol{c_s} = \sum_{i=1}^{T} \boldsymbol{k_{si}} \, a_{si} \quad (4)$$

An equivalent symmetric procedure is used to compute attention on $\boldsymbol{K_t}$ so as to obtain $\boldsymbol{c_t}$. The output of this block are two embeddings of size 50, as in our previous architecture.

Our method resembles the approach of Chen et al. [36], but with two important differences. First of all, they use *fine-grained co-attention* [28] instead of *coarse-grained*, so they consider each element of a sentence with respect to each element of the other sentence, leading to a higher computational footprint. The second difference is that they use *multiplicative attention* instead of the *additive* one: while the former is more indicated for tasks where it is important to consider the similarities between two inputs (as in the agreement inference task) the latter is more suitable for tasks where representations of relevant elements are unavailable [28], as in component classification and link prediction.

The resulting architecture has about 140,000 trainable parameters. If compared with other state-of-the-art neural architectures, such as $\text{BERT}_{\text{BASE}}$ and its 110 M parameters, ResAttArg is considerably smaller, and accordingly it is less computationally demanding.

### E. Optimization Model and Ensemble Learning

We consider a multi-task formulation for our learning problem. The loss function is given by the weighted sum of four different components: the categorical cross-entropy on three labels (source and target categories, link relation category) and an $L_2$ regularization on the network parameters.

Since the training of neural models is non-deterministic, the results of a single training procedure are influenced by the random seed that is used, thus they may not be reliable or reproducible [71], [72]. Such problem also affects our previous results [15], since they were obtained from a single training experiment.

We have decided to replicate that experiment by repeating the training procedure 10 times, with different seeds, obtaining 10 trained neural networks for each configuration. We will evaluate our models in two different ways. At first, we will consider the average of the scores obtained by every single network for each metric. Then, we evaluate the predictions obtained using all the 10 models in ensemble voting.

In our ensemble setting the class of each entity is assigned as the class voted by the majority of the networks. This technique is similar to the concept of bootstrap aggregating, also known as bagging [73]. However, while in standard bagging each model is trained on a random sample of the training set, here we train all the models on the same training set, since stochastic elements are already present in the training procedure itself. Indeed, the training process does involve non-deterministics steps, such as the initialization of the networks' weights, the selection of the elements for each batch, and the application of dropout. We have chosen this ensemble method for the sake of simplicity, but more advanced techniques do exist and may yield better results [74].

## IV. Corpora

We validate our approach on five corpora differing from each other in various dimensions: the domain of the documents, their average length, the formatting, and the argumentative model followed for the annotations.

### A. CDCP

The Cornell eRulemaking Corpus (CDCP) [23] consists of user-generated documents in which specific regulations are discussed. The authors have collected user comments from an eRulemaking website on the topic of Consumer Debt Collection Practices rule. The corpus contains 731 user comments, for a total of about 4,700 components, all considered to be argumentative.

As typical of user-generated data, the comments are not structured, and often present grammatical errors, typos, and do not follow usual writing conventions (such as the blank space after the period mark). This complicates pre-processing, since most of the off-the-shelf tools turn out to be inaccurate even in simple tasks such as tokenization.

Annotations follow the argument model proposed in [75], where links are constrained to form directed graphs. The corpus is suitable both for component and relation classification, since it presents 5 classes of propositions and two types of links. We will use the version of CDCP without nested proposition and guaranteed transitive closure.

Components are addressed as propositions, and they consist of a sentence or a clause. Propositions are divided into POLICY (815), VALUE (2160), FACT (746), TESTIMONY (1026), and REFERENCE (32). Only 3% of more than 43,000 possible proposition pairs are linked; almost all links are labeled as REASON (1,292), whereas only a few are labeled as EVIDENCE (46).

The unstructured nature of documents, the strong unbalance between the classes, and the presence of noise make the corpus particularly challenging for all the subtasks of argument mining, especially those that involve the relationships between components.

### B. AbstRCT

The AbstRCT Corpus [48] consists of abstracts of scientific papers regarding randomized control trials for the treatment of specific diseases (i.e., neoplasm, glaucoma, hypertension, hepatitis b, diabetes). The final corpus contains 659 abstracts, for a total of about 4,000 argumentative components. AbstRCT is divided into three parts: neoplasm, glaucoma, and mixed. The first one contains 500 abstracts about neoplasm, divided into train (350), test (100), and validation (50) splits. The remaining two are designed to be test sets. The glaucoma part contains 100 abstracts for that disease, the mixed one contains 20 abstracts for each disease.[4]

Components are labeled as EVIDENCE (2,808) and CLAIM (1,390), while relations are labeled as SUPPORT (2,259) and ATTACK (342).[5] About 10% of about 25,000 possible component pairs have a labeled relationship. The argumentative model chosen for annotation enforces only one constraint: claims can have an outgoing link only to other claims.

With respect of CDCP, this corpus is less noisy and the distribution of the classes is more balanced. We have chosen this as a benchmark to demonstrate that our approach is independent of the domain and of the argument model.

### C. DrInventor

The Dr. Inventor Argumentative Corpus (DrInventor) [76] is the result of an extension of the Dr. Inventor corpus [77], which includes an annotation layer containing argumentative components and relations. DrInventor consists of 40 scientific publications from computer graphics, which contain about 12,000 argumentative component labels, as well as annotations for other tasks.

The classes of argumentative components are DATA (4,093), OWN CLAIM (5,445), and BACKGROUND CLAIM (2,751). The former two are related to the concepts of premises and claims, while the latter is something in between, since it is a claim related to some background knowledge, such as that made by another author in a previous work. The relation classes are SUPPORTS (5,790), CONTRADICTS (696), and SEMANTICALLY SAME (44), since it is common practice in scientific publications to re-iterate the same claim (or more rarely the same data) multiple times.

Since DrInventor includes documents where the structure of the discourse is complex, and data are often presented along with claims, it makes argument mining more challenging: in more than 1,000 cases some components are split into multiple text sequences, located in non-contiguous parts of the documents. This phenomenon mostly concerns claims, but data are affected too, in fewer cases. This introduces the difficulty of recognizing different segments of the documents as part of a single component and makes link prediction more difficult to address through non-pipeline approaches.

The unbalanced distribution between the three classes and the presence of split components makes this corpus quite challenging for link prediction, a difficulty which is highlighted also by the low inter-annotator agreement reported in the original article.

### D. SciDTB

The SciDTB Argumentative Corpus [22] consists of 60 scientific abstracts from the ACL anthology, for a total of 353 argumentative components. Components can span across multiple sentences and can belong to six classes: PROPOSAL (110), ASSERTION (88), RESULT (64), OBSERVATION (11), MEANS (63), DESCRIPTION (7). The annotation scheme impose that each component can be linked only to another one, and presents only one class of argumentative relationship: SUPPORT.[6] Out of the 1884 possible pairs of components, only 126 (6.69%) are linked together by a SUPPORT. The challenging aspects in this corpus are its small size, which allows us to test our method on a low-resource setting, and its unbalance in the distribution of component classes.

### E. UKP-PE

The Persuasive Essays Corpus (UKP-PE) [4] consists of 402 documents from an online community where users post essays and other material, provide feedback, and advise each other. The dataset is divided into a test split of 80 essays and a training split with the remaining documents.

UKP-PE defines three classes of argumentative components: MAJOR CLAIM (751), CLAIM (1,506), and PREMISE (3,832). Premises may be linked to CLAIMS through relations of SUPPORT (3,613) or ATTACK (219). MAJOR CLAIMS are not linked to other components.[7] The classes of argument components are similar to those in other datasets. However, what distinguishes the UKP-PE corpus from the others is a more regular argumentation model, which is specific to this corpus alone. All argument graphs are trees. All roots are claims. All tree components belong to the same paragraph. Each premise has exactly one outgoing relation. Claims do not have outgoing relationships, they can only be supported/attacked by premises. The structure of the argumentation is also fairly regular. For example, major claims are usually present in the introduction or conclusion of an essay, and they are often the only argumentative component in the paragraph.

Thanks to the highly regular nature of the UKP-PE data, strong baselines heavily rely on document structure, like the position of the sentence in the essay, or whether a component is in the introduction or conclusion, or in the first or the last sentence of a paragraph [78]. At the same time, including such highly regular data in our analysis enables us to gain further insights on the strengths and limitations of a structure-agnostic approach like ours.

## V. Experimental Setting

We initially evaluate our new architecture against our previous model and the structured learning approach of [23] on CDCP, presenting an ablation study of the new components we have introduced. Then, we extend the evaluation to other four data sets, for which we compare our approach against the state-of-the-art. In our approach each component is involved in many pairs, both as a source and as a target, and accordingly it is classified

[4]Glaucoma and neoplasm documents of the mixed set are present also in the respective test set.

[5]The corpus allows also the distinction between CLAIM/MAJOR CLAIM and ATTACK/PARTIAL ATTACK. For the sake of consistency with previous works, this detail will not be considered.

[6]Other classes of relationship are present but they are considered not argumentative, therefore we do not consider them in our work.

[7]In fact, each component is linked to a MAJOR CLAIM via an attribute called *stance*. Therefore, one could use this dataset for stance detection, by creating explicit relationships toward the major claims. However, that would be outside the scope of this work.

TABLE I
CORPORA STATISTICS

| Corpus | Documents | Components | Component Pairs | Links |
|---|---|---|---|---|
| CDCP | 731 | 4,779 | 43,384 | 1,338 |
| AbstRCT | 659 | 4,198 | 25,938 | 2,601 |
| DrInventor | 263 | 13,591 | 243,266 | 8,705 |
| SciDTB | 60 | 353 | 1,884 | 126 |
| UKP-PE | 1,764 | 6,089 | 22,172 | 3,832 |

For UKP-PE we consider paragraphs containing arguments as "documents" and we do not include "self pairs" in the count of component pairs.

multiple times by the same network. The label will be assigned by the model by considering the average probability computed by the ensemble for each class, and by thus choosing the class with the highest score. Alternative approaches could be to assign the class that results to be the most probable in most of the cases, thus relying on a majority vote. A further option could be to simply consider the label with the highest confidence. However, the latter procedure might be more sensitive to outliers, because the misclassification of a component in just one pair would lead to the final misclassification of the component, regardless of all the other pairs. A deeper analysis of different techniques to address these issues is left to future research.

### A. Data Preparation

Table I report summary statistics of the datasets we use. Our architecture allows us to use the CDCP, AbstRCT, and SciDTB datasets directly, without need for further pre-processing.

For what concerns DrInventor, instead, specific data pre-processing is needed to address two aspects of this dataset: the presence of lengthy documents and split components. Lengthy documents make it inconvenient to consider all the possible pairs of argumentative units. Doing so would not only be infeasibile with regular computational resources, but it would also yield an extremely unbalanced dataset for link prediction, with less than 1% of pairs linked. We thus filtered out all the pairs that did not appear in the same section of the document, and whose argumentative distance is not included between −10 and +10. A second peculiarity of this dataset is the presence of components that include non-argumentative material. These "split components" are made of two sequences $x$ and $y$ separated by a third, non-argumentative sequence $z$. In those cases, we split $x$ and $y$ into two unrelated components, and attributed them the same label, the same links, and the same argumentative relations with the other components. The resulting dataset consists of about 8,700 links out of 240,000 possible pairs, which amount roughly to 3.6%. Among these links, SUPPORTS amount to 89%, CONTRADICTS to 10%, and the remaining 1% are SEMANTICALLY SAME relations.

Regarding UKP-PE, like others did before us [4], [23], we also consider exclusively pairs of components that belong to the same paragraph. However, many paragraphs contain only a single component. That is the case, for instance, with about 400 paragraphs containing a single major claim. In order to include also them in our pair-based classification method, we decided to introduce "self pairs" into our dataset, which are instances where the same component acts both as source and target. This significantly increases the number of pairs (from 22,000 to 28,000). So, to improve optimization and enable a comparison with previous approaches, we did not consider these pairs for link prediction and relation classification in validation and testing.

### B. Comparison With Other Methods

Not all the approaches to AM are easily compared against one another. This is the case, for example, of approaches that perform only few tasks versus end-to-end systems, or pipeline versus joint learning approaches. Since we perform component classification on propositions or sentences, to make our results comparable with architectures that perform it token-wise, we split each classified component into tokens that share the same label, and compute the evaluation of token-wise classification. Since the tokenization method may not be the same one used by other approaches, the final results may not be perfectly comparable, but we believe that this minor difference will not introduce appreciable errors.

We shall also remark that in our approach we consider argumentative components as already selected and perfectly bounded, therefore we perform component classification only between argumentative classes and we do not consider the "non-argumentative" class as a possibility. This makes our figures incomparable against those obtained by architectures that address both component identification and classification at once, such as [48], since they include "non-argumentative" among the possible classes and thus address a harder problem. A similar consideration holds regarding the pipeline approaches that perform evaluation of each step based on the result of the previous one instead of using the gold standard. In this case, the errors introduced by early steps introduce noise which may affect the evaluation of subsequent steps. It is once again the case of [48], where errors obtained during the first step may introduce noise in the link prediction/relation classification tasks, even if the authors report that such an error is neglegible. We could not find a solution to this problems, but we argue that, nevertheless, a qualitative evaluation of our method can still benefit from a comparison with these other approaches.

Due to these difficulties, in most of cases adapting existing techniques to a new corpus would be a very demanding task. For this reason, we compare our approach only against approaches that have already been tried on the same corpus.

### C. Optimization

For each corpus, we train the models on the corresponding training split. Since each corpus is characterized by different classes for component and relation classification, it would be impossible to test a model trained on a different corpus. Nonetheless, it would be possible to use the same model for the task of link prediction.

We shall remark that the hyper-parameters of the architecture and of the learning model have been tuned on the validation set of CDCP. It is also important to highlight that we use the same set of hyper-parameters in all the experiments. Our purpose

TABLE II
RESULTS OF THE ARGUMENT MINING ON CDCP

| Approach | Task Average | Link | Components | | | | | | Relation | | | |
|---|---|---|---|---|---|---|---|---|---|---|---|---|
| | | | **Macro** | VALUE | POLICY | TEST. | FACT | REF. | **Macro** | REASON | EVID. | NONE |
| Structured [23] | | | | | | | | | | | | |
| SVM | 50.0 | 26.7 | 73.5 | 76.4 | 77.3 | 71.7 | 42.5 | 100.0 | - | - | - | - |
| RNN | 43.5 | 14.6 | 72.7 | 73.7 | 76.8 | 75.8 | 42.2 | 100.0 | - | - | - | - |
| TSP+PLBA [35] | 56.5 | 34.0 | 78.9 | 80.7 | 83.3 | 79.0 | 51.6 | 100.0 | - | - | - | - |
| + checkpoint ens. | 56.7 | 33.8 | 79.5 | - | - | - | - | - | - | - | - | - |
| Transition BERT [46] | **59.9** | **37.3** | **82.5** | 83.2 | 86.3 | 84.9 | 58.3 | 100.0 | - | - | - | 98.3 |
| BERT + SA [47] | 52 | 25 | 81 | - | - | - | - | - | - | - | - | - |
| ResArg | | | | | | | | | | | | |
| Single [15] | 47.3 | 29.3 | 65.3 | 72.2 | 74.4 | 72.9 | 40.3 | 66.7 | - | 30.0 | 0.0 | - |
| Average | 47.8 | 25.0 | 70.5 | 72.3 | 75.4 | 73.5 | 41.4 | 90.0 | 41.8 | 25.1 | 2.5 | 97.8 |
| Ensemble | 52.1 | 28.8 | 75.5 | 75.4 | 79.6 | 76.3 | 46.4 | 100.0 | 42.3 | 28.6 | 0.0 | 98.2 |
| ResAttArg | | | | | | | | | | | | |
| Average | 51.57 | 27.40 | 75.75 | 77.84 | 80.09 | 76.42 | 44.39 | 100.00 | 42.69 | 27.88 | 2.22 | 98.03 |
| Ensemble | 54.22 | 29.73 | 78.71 | 80.37 | 82.55 | 81.19 | 49.42 | 100.00 | 42.95 | 30.56 | 0.00 | 98.32 |

From top to bottom: The best results of structured approaches based on SVM and RNN, two recent approaches, our previous result obtained with a single training of ResArg, the average scores of the same architecture trained 10 times, the scores of the ensemble learning setting of the same model, and finally the average and the ensemble scores of the new Attention-based architecture ResAttArg. When previous works do not report a score, we use the symbol "-".

is to test whether our approach can yield satisfactory results across different and heterogeneous corpora without the need of re-tuning, and therefore limiting its cost and its environmental impact [6]. Nonetheless, we are aware that performing a specific calibration for each corpus would probably improve our results.

We use the Adam optimizer [79] with parameters $b_1 = 0.9$ and $b_2 = 0.9999$, applying proportional decay of the initial learning rate $\alpha_0 = 5 \times 10^{-3}$. The weights of the four components of the loss function are set to 1 for the cross entropy of source and target, 10 for the cross entropy of relation, and $10^{-4}$ for the regularization. The training was early-stopped after 100 epochs with no improvements on the $F_1$ score of the Link class computed over validation data, except for DrInventor, where we early-stopped after 20 epochs of patience due to the dataset's size and much heavier computational footprint.

## VI. Results and Discussion

This section presents the experimental results on each corpus. To assess the contribution of the attention module, we compareResAttArg with ResArg. Moreover, we study the performance gain introduced by the ensemble approach. To be consistent with other works in this research field, we measure all the performances using the $F_1$ metric and report the values as percentages. For component and relation classification we consider the macro-averaged score. For link prediction, we consider the score of the positive class.

We structure the analysis of results on each corpus in separate subsections. Two more subsections are devoted to the analysis of computational costs and of the attention module.

### A. CDCP

We used the same validation set as in our earlier work [15], which was created by randomly selecting documents from the original training split with 10% probability. We used the remaining documents as training data and the original test split as is. To provide a summary evaluation, following [23], we measured the performance of the models by computing the $F_1$ score for links, propositions, and the average between the two. More specifically, for the links we measured the $F_1$ of the positive classes, whereas for the propositions we used the score of each class and then we computed the macro-average. We also reported the $F_1$ score for each relation class, alongside their macro-average. The NONE class of relation classification corresponds to the negative class of link prediction.

A first question we address was whether our results with a single model were solid or to what extent influenced by the non-deterministic nature of the training procedure. We compared our baseline model with the average scores obtained by 10 networks, with our ensemble setting, and against the structured approach used in [23]. The results are shown in Table II. The average computed over the 10 networks leads to a worse performance on Link prediction with respect to our previous results, which suggests that our previous results were due to a particularly "lucky" training. Nonetheless, the average score on the two tasks remains similar (between 47 and 48), just a few points below the state of the art. The ensemble approach substantially improves the results, outperforming the structured learning approach on both tasks. The results on link prediction are still below those obtained in the first experiment, if only by less than 1%.

Introducing the attention module in the architecture leads to appreciable improvements for both the average and the ensemble approach. In particular, the latter performance outperforms our previous result in all the three tasks. As far as relation label prediction, our approaches fail to predict the EVIDENCE relation. This is a negative result, but hardly surprising, since EVIDENCE is a rather rare class in this dataset (less than 1% of all relations). Also, we can see in Table X that the use of attention strongly reduces the amount of training epochs (−56%) as well as the standard deviation.

Our results on component classification are similar to the ones obtained by using TSP and PLBA [35], while Transition-based

| Structured SVM full | | Predicted P | F | T | V | R |
|---|---|---|---|---|---|---|
| True | P | 0.76 | 0.05 | 0.04 | 0.16 | 0.00 |
| | F | 0.04 | 0.44 | 0.10 | 0.42 | 0.00 |
| | T | 0.01 | 0.06 | 0.72 | 0.21 | 0.00 |
| | V | 0.05 | 0.11 | 0.08 | 0.76 | 0.00 |
| | R | 0.00 | 0.00 | 0.00 | 0.00 | 1.00 |

| ResArg Single | | Predicted P | F | T | V | R |
|---|---|---|---|---|---|---|
| True | P | 0.76 | 0.06 | 0.01 | 0.17 | 0.00 |
| | F | 0.06 | 0.42 | 0.08 | 0.44 | 0.00 |
| | T | 0.00 | 0.06 | 0.75 | 0.18 | 0.00 |
| | V | 0.07 | 0.12 | 0.10 | 0.70 | 0.00 |
| | R | 0.00 | 0.00 | 0.00 | 0.00 | 1.00 |

| ResArg Ensemble | | Predicted P | F | T | V | R |
|---|---|---|---|---|---|---|
| True | P | 0.78 | 0.07 | 0.01 | 0.14 | 0.00 |
| | F | 0.03 | 0.54 | 0.06 | 0.36 | 0.00 |
| | T | 0.00 | 0.07 | 0.77 | 0.15 | 0.00 |
| | V | 0.04 | 0.15 | 0.09 | 0.72 | 0.00 |
| | R | 0.00 | 0.00 | 0.00 | 0.00 | 1.00 |

| ResAttArg Ensemble | | Predicted P | F | T | V | R |
|---|---|---|---|---|---|---|
| True | P | 0.80 | 0.08 | 0.01 | 0.11 | 0.00 |
| | F | 0.02 | 0.52 | 0.05 | 0.41 | 0.00 |
| | T | 0.00 | 0.05 | 0.80 | 0.14 | 0.00 |
| | V | 0.04 | 0.10 | 0.06 | 0.80 | 0.00 |
| | R | 0.00 | 0.00 | 0.00 | 0.00 | 1.00 |

Fig. 3. Confusion matrices for component classification on CDCP. Columns and rows refer to the classes POLICY, FACT, TESTIMONY, VALUE, and REFERENCE, respectively. From left to right: Structured Learning approach, our previous result with ResArg [15], ResArg used in ensemble fashion, and ResAttArg used in ensemble fashion.

TABLE III
RESULTS OF COMPONENT CLASSIFICATION ON ABSTRCT

| Level | Approach / Test set | Neoplasm $f_1$ | Neoplasm $F_1$ | Neoplasm E | Neoplasm C | Glaucoma $f_1$ | Glaucoma $F_1$ | Glaucoma E | Glaucoma C | Mixed $f_1$ | Mixed $F_1$ | Mixed E | Mixed C |
|---|---|---|---|---|---|---|---|---|---|---|---|---|---|
| Token | Transformer-based [48] | | | | | | | | | | | | |
| | BioBERT+GRU+CRF | 90 | 84 | 90 | 87 | 92 | **91** | 91 | 93 | 92 | **91** | 92 | 91 |
| | SciBERT+GRU+CRF | 90 | 87 | 92 | 88 | 91 | 89 | 91 | 93 | 91 | 88 | 93 | 90 |
| | Our approach | | | | | | | | | | | | |
| | ResArg (avg) | 91 | 88 | 94 | 83 | 92 | 87 | 95 | 80 | 91 | 88 | 94 | 81 |
| | ResArg+Ensemble | 91 | 88 | 94 | 82 | 93 | 88 | 95 | 82 | 92 | 88 | 95 | 92 |
| | ResAttArg (avg) | 91 | 89 | 94 | 84 | 92 | 88 | 95 | 81 | 92 | 89 | 94 | 83 |
| | ResAttArg+Ensemble | **92** | **90** | 95 | 86 | **93** | 89 | 96 | 83 | **93** | 90 | 95 | 85 |
| Component | Our approach | | | | | | | | | | | | |
| | ResArg (avg) | 87 | 86 | 90 | 82 | 88 | 86 | 92 | 79 | 88 | 87 | 91 | 82 |
| | ResArg+Ensemble | 88 | 86 | 91 | 82 | 89 | 87 | 93 | 82 | 89 | 88 | 92 | 83 |
| | ResAttArg (avg) | 87 | 86 | 90 | 82 | 89 | 86 | 92 | 81 | 89 | 88 | 91 | 84 |
| | ResAttArg+Ensemble | **89** | **88** | 91 | 84 | **90** | **88** | 93 | 87 | **91** | **90** | 93 | 83 |

We report the $F_1$ score related to the micro average ($F_1$), the macro average ($F_1$), the EVIDENCE class (E), and the CLAIM class (C) obtained on the 3 test sets. Our approach is not directly comparable with component classification of [48] and the comparison must be considered qualitative.

BERT [46] greatly surpasses our approach in both tasks. It is worth remarking that both these approaches use a mixture of symbolic and subsymbolic features, and that BERT has about 1,000 times more trainable parameters than one of the networks in our ensemble. Bao et al. [46] report they fine-tuned their model 50 epochs with early stopping strategy. In our experiments, the average time required to train a BERT model for a single epoch is remarkably higher than the time required to train our models (more details on the computational cost will be given in Section VI-F).

To estimate the agreement among the networks in the ensemble architecture, and have a measure of the robustness against the implicit randomness of the training procedure, we have computed Krippendorff's alpha [80] for the three tasks. We obtained $\alpha = 0.70$ for component classification, and $\alpha = 0.44$ for both link prediction and relation classification. These values are similar to the IAA obtained by the authors of the corpus, and confirm the difficulty of the link prediction task.

Fig. 3 shows confusion matrices for component classification on CDCP. Unsurprisingly, the most common mistake regards the prediction of facts as values – VALUE being by far the largest class in the corpus, and so affected by many false positives. Such an ambiguity between the two classes has also been reported during the annotation process.

Interestingly, the confusion matrices of the structured approach and of our methods are quite similar. We speculate that our networks may have learned a behavior similar to that produced by the structured approach, with no need to receive any of the constraints or information regarding the argumentative structure that are instead injected in the structured approach.

TABLE IV
RESULTS OF RELATION CLASSIFICATION ON ABSTRCT

| Approach / Test set | Neoplasm | Glaucoma | Mixed |
|---|---|---|---|
| Transformer-based [48] | | | |
| BioBERT | 64 | 58 | 61 |
| SciBERT | 68 | 62 | 69 |
| SciBERT + Weighted Loss | 68 | **70** | **70** |
| RoBERTa | 67 | 66 | 67 |
| RoBERTa + Weighted Loss | 68 | 67 | 67 |
| Our approach | | | |
| ResArg (avg) | 59 | 57 | 60 |
| ResArg+Ensemble | 63 | 62 | 68 |
| ResAttArg (avg) | 66 | 63 | 63 |
| ResAttArg+Ensemble | **71** | 68 | 68 |

## B. AbstRCT

For what concerns AbstRCT, we compare our architectures against the best methods presented by its authors [48], whose results are reported in the first rows of Tables III and IV. We trained and validated our model on the respective splits of the Neoplasm dataset, using the remainder of the dataset for testing. For reasons we already explained, the approach presented by

TABLE V
RESULTS OF RESATTARG WITH ENSEMBLE FOR LINK PREDICTION AND RELATION CLASSIFICATION ON ABSTRCT

| Test set | **Link** | **Relation** | SUPPORT | ATTACK | NONE |
|---|---|---|---|---|---|
| Neoplasm | 54.43 | 70.92 | 52.77 | 65.38 | 94.54 |
| Glaucoma | 55.23 | 68.40 | 54.73 | 56.00 | 94.36 |
| Mixed | 51.20 | 67.66 | 49.62 | 59.09 | 94.21 |

TABLE VI
RESULTS OF COMPONENT CLASSIFICATION ON DRINVENTOR

| Level | Approach | **Macro** | O C | B C | D |
|---|---|---|---|---|---|
| Token | Bi-LSTM [21] | 44.70 | - | - | - |
| | RESARG (avg) | 58.27 | 72.31 | 51.36 | 51.14 |
| | RESARG+Ens | 61.16 | 75.77 | 54.24 | 53.48 |
| | RESATTARG (avg) | 61.77 | 73.66 | 57.70 | 53.95 |
| | RESATTARG+Ens | **65.71** | 78.03 | 61.58 | 57.53 |
| Component | RESARG (avg) | 60.62 | 65.65 | 45.63 | 70.56 |
| | RESARG+Ens | 62.97 | 68.97 | 48.03 | 71.90 |
| | RESATTARG (avg) | 66.19 | 68.61 | 56.07 | 73.89 |
| | RESATTARG+Ens | **69.64** | 73.13 | 59.63 | 76.15 |

We report the macro $F_1$ score, and the $F_1$ for the classes OWN CLAIM, BACKGROUND CLAIM, DATA.

TABLE VII
RESULTS OF RESATTARG WITH ENSEMBLE FOR LINK PREDICTION, AND RELATION CLASSIFICATION ON DRINVENTOR

| **Link** | **Relation** | SUPP | CON | SEM SAME | NONE |
|---|---|---|---|---|---|
| 43.66 | 37.72 | 45.90 | 6.61 | 0 | 98.37 |

Mayer et al. [48] is not directly comparable with ours, therefore the comparison can only be qualitative. To ease comparison with future approaches, we report in Table V some additional details on our results.

As for component classification, RESATTARG with ensemble yields the best result, performing comparably with the state of the art. Our approaches obtain substantially better scores for EVIDENCE than CLAIM on all datasets. Similarly to the Transformer-based approaches, our architectures perform better on the mixed test set than on the neoplasm one. We yield better results on all datasets for what concerns the micro $f_1$ score. However, for what concerns macro $F_1$, although our architecture improves the previous approaches on Neoplasm, it is outperfomed by BioBERT on Glaucoma and Mixed. In relation classification, RESATTARG with ensemble outperforms all the other models on Neoplasm, and it performs about 2% worse than the state of the art on Mixed and Glaucoma. It is interesting to notice that in this task BioBERT is largely outperformed by our approach. Almost all the metrics confirm that the introduction of attention and ensemble improve our architectures. The agreement between the networks RESATTARG is very high for token-wise component classification in each dataset ($0.81 \leq \alpha \leq 0.83$), and lower but still acceptable for the other two tasks ($\alpha = 0.67$ on neoplasm and $\alpha = 0.62$ for the other two). The introduction of attention has importantly reduced the amount of training epochs ($-29\%$) and the standard deviation. On this corpus, RESARG requires about half the amount of training epochs it required on CDCP, while RESATTARG requires a few less.

These good results indicate that our method may be a valuable approach with well-structured corpora. Moreover, such results are attained without resorting to contextual embeddings or pre-training on domain-related corpora, but by only relying on non-contextual, general-purpose embeddings.

## C. DrInventor

To the best of our knowledge, the only approach tested on this corpus is the architecture for token-wise component classification used by Lauscher et al. [21], which makes use of GloVe embeddings and a Bi-LSTM followed by a feed-forward neural network with a single hidden layer as classifier. We thus consider such an approach as a baseline. Like Lauscher et al., we reserved 30% of the documents of the DrInventor corpus as test set, and 20% of the remaining part as validation set. It is worth remarking that for the tasks of link prediction and relation classification we are considering a limited number of pairs.

Tables VI and VII includes a detailed report of our performance on the dataset. We outperform the baseline by a wide margin. Moreover, we address two additional tasks, link prediction and relation classification, thus offering a benchmark for future work. These results confirm once more that attention and ensemble together give a crucial contribution to the classifier.

Differently from previous experiments, the agreement between the networks RESATTARG is similar for all the tasks, with only $\alpha = 0.56$ for component classification and $\alpha = 0.60$ for the remaining tasks. The agreement for Component Classification is lower than on the previous datasets and may suggest that this dataset is more challenging. This is the only case where RESARG requires less training epochs than RESATTARG, but the difference is neglegible.

Our model is incapable of classifying the SEMANTICALLY SAME relation and has difficulties also with CONTRADICTS. That is hardly surprising, if we consider that these are the two least represented classes in this dataset. It is less straightforward to understand why the model is better at classifying BACKGROUND CLAIM rather than DATA, even if the latter are more represented than the former. We speculate it may be related to the fact that in some instances data may amount to citations or text other than proper sentences.

## D. SciDTB

Previous experiments on this dataset were conducted using BiLSTM and CRF, exploiting syntactical, positional, and discourse features [22]. The authors performed component classification at token level using BIO tagging and validated it through a 10-fold cross-validation setting. We have decided to pursue a different experimental setting: we randomly split the corpus into train, validation, and test folds with an approximate rate of 60%, 20%, 20%, imposing the constraint that each fold must contain at least 1 instance of each component class. We are aware that such decision makes our results impossible to compare with previous approaches, but we are deeply convinced that this is the best approach for the task at hand. Indeed, since some component classes are under-represented in the dataset (for

TABLE VIII
RESULTS OF AM ON SCIDTB

| Level | Approach | **Task Average** | Link **Yes** | Link No | Component **Macro** | Micro | PROP | ASS | RES | OBS | MEANS | DESC |
|---|---|---|---|---|---|---|---|---|---|---|---|---|
| Token | RESARG (avg) | 41.23 | 38.99 | 96.20 | 43.46 | 54.03 | 64.18 | 55.32 | 52.08 | 59.57 | 29.64 | 0.00 |
| | RESARG+Ensemble | 47.72 | 41.67 | 96.15 | 53.78 | 61.11 | 68.18 | 62.50 | 72.00 | 100.00 | 20.00 | 0.00 |
| | RESATTARG (avg) | 43.58 | 45.25 | 96.97 | 41.90 | 52.22 | 60.48 | 50.68 | 51.98 | 55.00 | 33.02 | 0.27 |
| | RESATTARG+Ensemble | **56.17** | **50.00** | 97.28 | **62.35** | 70.83 | 78.26 | 66.67 | 72.00 | 100.00 | 57.14 | 0.00 |
| Component | RESARG (avg) | 40.51 | 38.99 | 96.20 | 42.03 | 52.53 | 64.21 | 52.46 | 49.98 | 58.38 | 27.16 | 0.00 |
| | RESARG+Ensemble | 46.51 | 41.67 | 96.15 | 51.35 | 58.23 | 65.83 | 58.80 | 68.06 | 100.00 | 15.42 | 0.00 |
| | RESATTARG (avg) | 43.05 | 45.25 | 96.97 | 40.84 | 51.07 | 59.94 | 48.27 | 51.18 | 55.56 | 29.92 | 0.21 |
| | RESATTARG+Ensemble | **55.67** | **50.00** | 97.28 | **61.33** | 70.16 | 78.67 | 66.10 | 70.47 | 100.00 | 52.74 | 0.00 |

For link prediction, we report the $F_1$ score related to the two classes.

TABLE IX
RESULTS OF AM ON UKP-PE

| Approach | **Task Average** | **Link** | Component **Macro** | P | C | M C | Relation **Macro** | SUPPORT | ATTACK | NONE |
|---|---|---|---|---|---|---|---|---|---|---|
| ILP joint model | **70.6** | 58.5 | **82.6** | 90.3 | 68.2 | 89.1 | - | - | - | 91.8 |
| structured SVM | 68.9 | **60.1** | 77.6 | 90.3 | 64.5 | 80.0 | - | - | - | - |
| structured RNN | 64.8 | 50.4 | 79.3 | 87.6 | 62.0 | 88.3 | - | - | - | - |
| RESARG (avg) | 37.0 | 29.9 | 44.1 | 78.5 | 41.2 | 12.6 | 38,2 | 29.7 | 1.1 | 83.9 |
| RESARG Ensemble | 37.1 | 30.4 | 43.9 | 80.8 | 40.2 | 10.5 | 38.3 | 30.8 | 0.0 | 84.2 |
| RESATTARG (avg) | 43.3 | 33.5 | 53.0 | 78.5 | 42.0 | 38.7 | 40.0 | 32.9 | 0.0 | 87.0 |
| RESATTARG Ensemble | 44.4 | 36.3 | 52.5 | 81.6 | 42.1 | 33.9 | 41.6 | 36.1 | 0.0 | 88.6 |

For link prediction, we report the $F1$ related to the positive class. The score of the negative class corresponds to the score obtained by the NONE class in the task of relation classification. When previous works do not report a score, we use the symbol "-".

example, there are only 11 instances of OBSERVATION and 7 of DESCRIPTION), some folds will not contain any instances of them, and therefore some of the tests will completely ignore those classes, resulting in unreliable measures. For what concerns link prediction, previous experiments were conducted also considering non argumentative relationships, so it is impossible to compare those results to ours.

Due to the small size of the corpus, the architectures overfitted on both the task of component classification and link prediction, obtaining the perfect score on the training set. The results on the test set are showed in Table VIII. The performance on component classification are comparable to the measures obtained on DrInventor, and the architecture clearly have difficulty to recognize MEANS and DESCRIPTION. Surprisingly, OBSERVATION, the second least represented class, is always correctly classified. While the use of ensemble plays a key role, once again it is the combination of attention and ensemble that leads to the best result. The agreement between the networks is extremely low ($\alpha < 0.30$), and the measure is deeply affected by the error on the DESCRIPTION class. For what concerns link prediction, similar considerations can be drawn, except that the agreement is considerably higher, but still unreliable ($0.43 < \alpha < 0.46$). The number of necessary training epochs is higher than in previous corpora for both the models. Once more, RESATTARG requires less epochs than RESARG, but the standard deviation for both models is similarly high.

## E. UKP-PE

UKP-PE comes with two strong baselines: the ILP joint model proposed by the authors of the dataset [4] and the structured learning approach by Niculae et al. [23]. They heavily rely on corpus-specific structural features. For example, it is possible to obtain a 47.7 F1 score for major claim detection by only using structural information [78]: something that has never been even attempted in other datasets, because it would not make sense to do so. We compare our results based on the original test split of the dataset, using about 10% of the documents of the training split as a validation split. As shown in Table IX, our approach is largely below the baselines, with a difference in $F_1$ scores between 20% and 30%. Both the use of attention and ensemble bring consistent improvements with respect to the base model. The agreement between the networks is also low, with $\alpha = 0.57$ for component classification and $\alpha = 0.38$ for link prediction, assessing them as nearly acceptable for the first task but unreliable for the others. The number of training epochs required is the highest for both models and also the standard deviation is very large. This suggests that the training process is probably unstable and it is difficult to make the model converge. Investigating the type of errors, we see that most CLAIMS are predicted as PREMISES and that most MAJOR CLAIMS are predicted as CLAIMS. As previously noted, failure to outperform the baselines by ignoring dataset-specific structural knowledge is not suprising, but it gives us a valuable indication of the limits of a structure-agnostic approach like ours.

## F. Analysis of Computational Costs

We shall now turn to the analysis of the computational cost. Table X reports the average number of epochs required to train our models on each dataset (we do not include the epochs of patience during which there was no improvement on the validation set). Table XI reports the seconds spent by one epoch of training on a single GPU GeForce GTX 1080 Ti. For comparison, we also report the average time required to train a BERT model [37][8] on the same hardware. We considered two typical training regimes: fine-tuning of the entire model (*BERT fine-tuned*), and training of the classification head only (*BERT freezed*). Training BERT freezed for one epoch requires ten times as long as training RESATTARG for one epoch. Training BERT fine-tuned for one epoch requires thirty times as long. The total cost of training will depend on the epochs required to reach the result. Standard practice when using BERT models in AM is to fine-tune for just 3 epochs (see, e.g., [48] and references therein). However, 3 epochs is a lower bound. For instance, the best result on the CDCP dataset so far have been obtained by fine-tuning a BERT-based model for 50 epochs with early stopping strategy [46]. These results could not be outperformed by training a plain BERT model for component classification for 3 epochs only [47] Moreover, transformer models like BERT often are but the backbone of more sophisticated AM architectures. Clearly, a backbone with such a large computational cost per single epoch greatly limits the possibility to test different architectures or to perform model selection and hyperparameter tuning. In conclusion, training a single RESATTARG architecture requires less time than fine-tuning a BERT model for 3 epochs (2856 vs 3399 seconds), let alone 50 epochs. Moreover, an ensemble of smaller architectures could better exploit parallel training and, importantly, can have a much lower computational footprint at inference time (see Section III-D: 110 M parameters for a plain BERT model vs 1.4 M parameters for an ensemble of 10 RESATTARG models).

TABLE X
AVERAGE NUMBER OF EPOCHS REQUIRED FOR TRAINING FOR EACH DATASET AND STANDARD DEVIATION

| | CDCP | AbstRCT | DrInventor | SciDTB | UKP-PE |
|---|---|---|---|---|---|
| RESARG | 190±95 | 105±73 | 29±11 | 202±80 | 212±232 |
| RESATTARG | 84±39 | 75±45 | 33±7 | 159±87 | 183±128 |
| Difference | -106 | -30 | +3 | -43 | -29 |
| Difference as % | -56% | -29% | +11% | -21% | -14% |

TABLE XI
AVERAGE NUMBER OF SECONDS REQUIRED FOR EACH TRAINING EPOCH

| | CDCP | AbstRCT | DrInventor | SciDTB | UKP-PE |
|---|---|---|---|---|---|
| BERT fine-tuned | 1,133 | 864 | 4,240 | 37 | 334 |
| BERT freezed | 421 | 319 | 1,535 | 14 | 122 |
| RESARG | 8 | 4 | 20 | <1 | 3 |
| RESATTARG | 34 | 19 | 62 | 1 | 10 |

## G. Analysis of Attention

The attention module, besides improving the overall performance of our architecture, can also be used to enhance the interpretability of the underlying neural model. To this aim, the normalized weights $a_s$ attributed by the module to each token are typically used as indicators of which words does the model consider important for the tasks. Indeed, whether this mechanism is really helpful for improving the explainability of neural models is still a matter of discussion [28], [81], [82]. Fig. 4 provides some visualizations of such scores.

[8] We used the `bert-base-uncased` model.

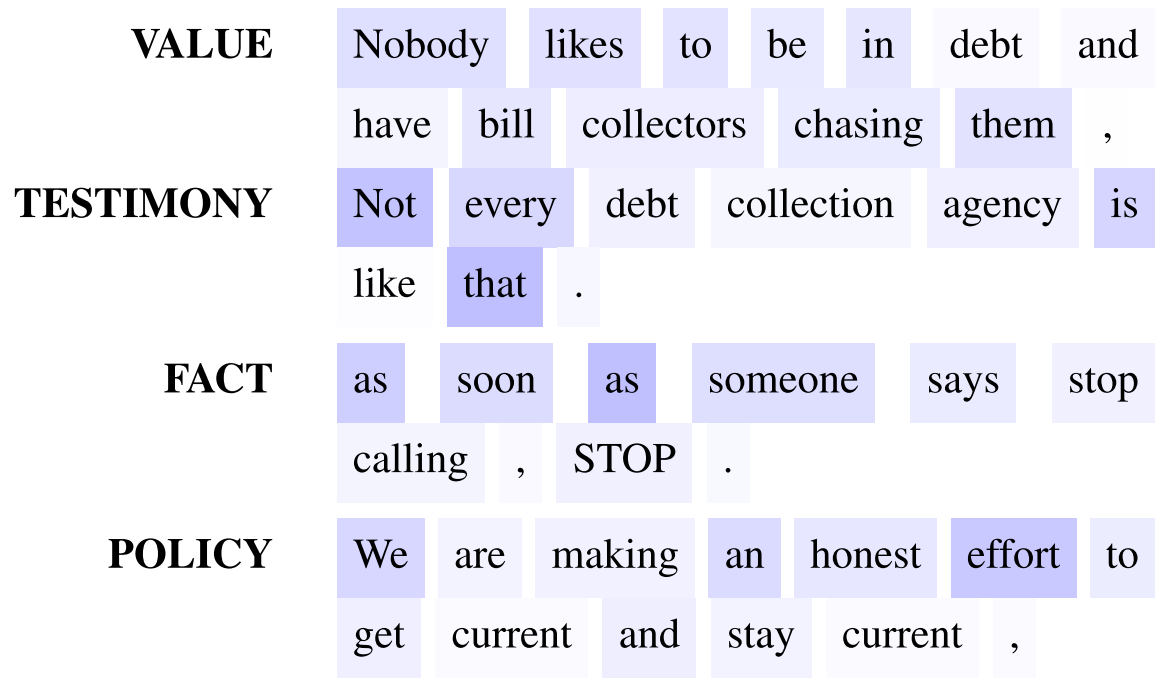

Fig. 4. Visualization of the attention score given to some sentences, along with their predicted class. Words in blue are the ones that receive more attention.

TABLE XII
TOP 20 TOKENS OF CDCP BY AVERAGE ATTENTION

| General Case | | Link Case | |
|---|---|---|---|
| Word | Score | Word | Score |
| ?! | 59 | Please | 69 |
| Please | 46 | Surely | 36 |
| Seriously | 41 | thats | 34 |
| professionals | 37 | phased | 34 |
| Nobody | 36 | Depending | 32 |
| Voice | 33 | Owing | 30 |
| Make | 31 | Continuing | 30 |
| phased | 31 | relentless | 29 |
| Enough | 29 | provider | 29 |
| mislead | 29 | spam | 28 |
| Calling | 29 | danger | 28 |
| fashioned | 29 | whenever | 27 |
| thats | 29 | Luckily | 27 |
| Not | 29 | refinancing | 27 |
| Unethical | 28 | massive | 26 |
| Everything | 27 | This | 25 |
| Two | 27 | hung | 25 |
| stone | 27 | finds | 25 |
| particularly | 26 | Express | 24 |
| nor | 26 | Emails | 24 |

In an attempt to assess the impact of attention on interpretability, we analyzed RESATTARG by measuring which are the tokens that, on average, receive the most attention. Tables XII and XIII report the 20 tokens from the CDCP and SciDTB corpora, respectively, that receive on average the most attention both in the general case (leftmost part in the tables), and in case the network predicts a link (rightmost part).

This kind of analysis gives mixed results. In the general case, tokens should give us hints about which elements are used by the network to distinguish between different components. The CDCP corpus stands out since in the first position there is the punctuation sign "?!". This sign as well as other tokens seem good indicators of emotional involvement (e.g., also "please", "seriously"). We speculate that they may be useful for the network to discriminate objective components from subjective ones (e.g., TESTIMONIES from VALUES, CLAIMS from EVIDENCES). These results seem to confirm previous findings, regarding the importance of subjectivity analysis [83] for the task of argument mining [84]. For what concerns link prediction, among the top-ranked tokens some discourse markers can be observed (e.g., "depending", "owing", "whenever", "firstly", "second", "while", "even"), as well as verbs in present continuous form, but also words that do not seem correlated with the task, such as "spam" or "phased". These results provide an additional example of the ambivalence of the study of attention weights to obtain an explanation of classification performed by black-box models.

TABLE XIII
TOP 20 TOKENS OF SCIDTB BY AVERAGE ATTENTION

| General Case | | Link Case | |
|---|---|---|---|
| Word | Score | Word | Score |
| selectional | 13 | Performance | 23 |
| However | 12 | However | 13 |
| comparison | 12 | Firstly | 12 |
| word-based | 12 | Unlike | 12 |
| State-of-the-art | 11 | word-based | 11 |
| different | 11 | one | 11 |
| Second | 11 | literature | 10 |
| one | 10 | linguistics | 10 |
| This | 10 | arguments | 10 |
| arguments | 10 | themselves | 10 |
| Firstly | 10 | Unfortunately | 10 |
| compare | 10 | Deceptive | 10 |
| lexico | 10 | Experiments | 10 |
| propose | 10 | The | 10 |
| Rich | 10 | Even | 10 |
| comparatively | 10 | comparatively | 10 |
| Even | 10 | Compared | 10 |
| While | 10 | advantages | 9 |
| straight-forward | 10 | improve | 9 |
| improve | 10 | detection | 9 |

## VII. CONCLUSION

In this article we presented RESATTARG, a new neural architecture for argument mining based on residual networks, multi-task learning, neural attention, and ensemble learning. Our approach does not rely on dataset-specific architectural choices such as structural features or encodings. On the contrary, it only uses general-purpose embeddings and a broadly applicable distance feature, making it suitable for any domain and argumentative model. Moreover, RESATTARG is considerably smaller than other state-of-the-art approaches, making it less expensive to train, and more sustainable from an environmental perspective [6], [7].

In spite of its lower computational footprint, RESATTARG equals or outperforms state-of-the-art architectures on a variety of tasks and datasets, with a notable exception of a dataset whose structural properties are crucial for a correct identification of argumentative components. We conducted ablation studies to evaluate the performance gain yielded by the attention module and the ensemble learning addition and to compare to our previous work [15]. The use of ensemble also increases the robustness of this approach against the intrinsic randomness of neural architecture training. The attention module could be instrumental to interpreting the behavior of the model, although our analysis gives mixed results on that.

The main limitations of RESATTARG are its limited scalability to large documents, and its failure to accommodate task-specific structural constraints. The latter is a design choice, and our results show that highly regular datasets and tasks are best address with dedicated architectures. Alternatively, neural-symbolic approaches [24], [85] may enable a systematic and modular integration of background knowledge. Such a knowledge would contribute during the optimization process, so as to influence and improve the training, without compromising the generality of the neural architecture. However, they risk exasperating the existing scalability issues [86]. When facing very high numbers of argument pairs, we addressed scalability by limiting the range of argumentative relationships using a fixed-size window, which, although cognitively plausible, and in agreement with annotated dataset statistics, meant imposing an additional constraint on the model of the argument. Alternatives to our pair-based approach include multiple-choice classifiers [48], pointer networks [34], and sequence labelling [18]. Such methods should scale better, but they enforce a constraint on the argument model as well, imposing that any component can have only one outgoing relationship, which makes them unsuitable to some corpora. Finally, since all the datasets have a strong unbalance, the use of weighted loss [48] or augmentation techniques [87], [88] may contribute further performance gain.

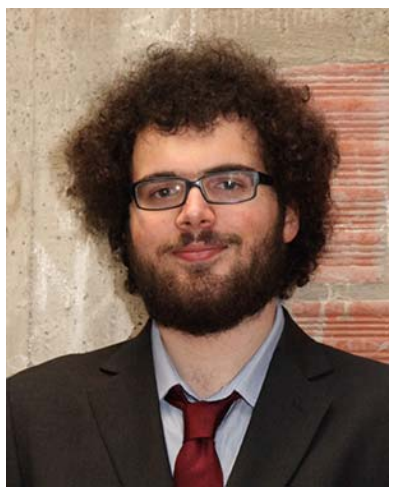

**Andrea Galassi** is currently a junior Assistant Professor with the Department of Computer Science and Engineering, University of Bologna, Bologna, Italy. His research interests include artificial intelligence and machine learning, focusing on argumentation mining and related natural language processing tasks.

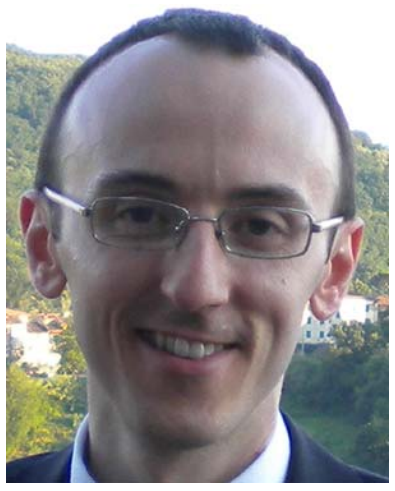

**Marco Lippi** is currently an Associate Professor with the Department of Sciences and Methods for Engineering, University of Modena and Reggio Emilia, Modena, Italy. He held positions with the Universities of Florence, Siena and Bologna, Italy, and he was a Visiting Scholar with Sorbonne University Pierre and Marie Curie Campus, Paris, France. His research interests include machine learning and artificial intelligence, with applications to several areas, including argumentation mining, legal informatics, and medicine. In 2012 he was the recipient of the E. Caianiello Prize for the best Italian Ph.D. Thesis in the field of neural networks.

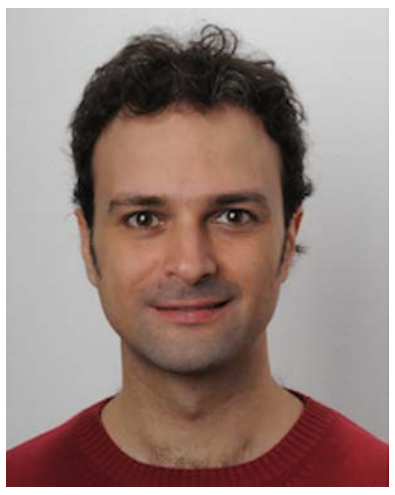

**Paolo Torroni** has been an Associate Professor with the University of Bologna, Bologna, Italy, since 2015. His main research focuses on artificial intelligence. He edited more than 20 books and special issues and authored more than 150 articles in computational logics, multi-agent systems, argumentation, and natural language processing.

Open Access funding provided by 'Alma Mater Studiorum - Università di Bologna' within the CRUI CARE Agreement